# Kunchenko's Polynomials for Template Matching

Oleg Chertov, Taras Slipets
Applied Mathematics Department
National Technical University of Ukraine "Kyiv Polytechnic Institute"
Kyiv, Ukraine
chertov@i.ua, taras.slipets@gmail.com

*Abstract*—This paper reviews Kunchenko's polynomials using as template matching method to recognize template in one-dimensional input signal. Kunchenko's polynomials method is compared with classical methods – cross-correlation and sum of squared differences according to numerical statistical example.

## I. INTRODUCION

Recognition of specified objects in data sets is well-known problem in various industries: from computer vision and handwriting characters recognition to data mining. Such technique is called template matching [1].

The problem statement of matching data with template includes input data sets to analyse, specified template and sometimes search domain. It's very important to represent available input data in appropriate form and define matching criteria.

In this paper, we consider 1D digital signal. Although this approach can be used with 2D images because several techniques to transform 2D images into one-dimensional representation have been proposed recently [2].

## II. RELATED WORK

There are two very different approaches to find templates in available data set [3].

The first one is to use artificial neural networks after learning phase [4]. In terms of mathematics, neural network's learning is a multi-parametric nonlinear optimization problem. A large number of right answers are required and that is not always possible.

The second approach consists in comparing either special features of template with data set (feature-based matching) or entire template with part of data set (template-based matching) [3]. As far as we consider one-variable function analysis, we don't focus on feature-based template matching, for instance angles or edges at pictures [5, 6]. Let's take a look at technique of entire template search in input data set.

This method is one of the first in history. In this sort of problems when comparing template with signal Euclidean metric, sum of squared differences, cross-correlation and several others are used [7-9]. However, we can try to solve this problem in another way – we may approximate input signal using polynomial template-based function [10]. In this case, areas of the most close signal and polynomial approximation that are over some threshold value can be considered as places of signal where template occurred.

It is well known that a Taylor polynomial is a polynomial to approximate a function the best way in a neighborhood of a given point. Though, the approximated function must have derivatives of appropriate order in this surrounding. Conditions that allow to approximate function with Fourier series over linearly independent orthogonal functions system that made up basis of considering space with inner product are much weaker. But from the template matching point of view Kunchenko's polynomials using seems to be very prospective [11, 12]. It is a result of the space where the polynomials are built. Such space is a subspace of Euclidean or Hilbert space that has particular element called generative element. Given template can be considered as generative element. After that template matching problem can be discussed as simpler problem of finding closest Kunchenko's approximation polynomial using some metrics [10].

## III. THEORETIC BACKGROUND

Let us assume function $f(x)$ called generative function and defined on the interval $[a, d]$. After that we denote set of ordered generated functions as follows:

$$u_v(x) = \varphi_v[f(x)], \; x \in [a,d], \quad (1)$$

where $\varphi_v(\cdot)$ – real functions fitted in some particular way.

In case of defining linear operations – addition and multiplication by number – set of all generated functions' linear combinations will make linear space.

Since some of generated functions (1) can be linear dependent, let us define space-constructing set of functions. This set is constructed as union of linear independent generated functions from (1). Thus, we'll call linear space over such set as linear space over independent generated functions (linear Kunchenko's space) and denote LFKu.

Now let us define inner product of two elements $u_v(x)$ and $u_k(x)$ in LFKu space as usual and call this product as correlant of these two elements:

$$\Psi_{v,k} \equiv (u_v(x), u_k(x)) \equiv \int_a^d \varphi_v[f(x)] \cdot \varphi_k[f(x)]dx. \quad (2)$$

According to (2) we can define distance $\rho_{vk}$ between two elements $u_v(x)$ and $u_k(x)$ from Kunchenko's linear space as norm of difference between two functions:

$$\rho_{vk}^2 \equiv \|u_v(x) - u_k(x)\|^2 = \int_a^d [u_v(x) - u_k(x)]^2 dx. \quad (3)$$

LFKu space where set of generated functions consists only of linear independent and pairwise (totally or partly) nonorthogonal functions we'll call Kunchenko's space with generative element. Under partially nonorthogonal generated functions we consider such set of functions where some of them are pairwise nonorthogonal and the other part of functions is pairwise orthogonal.

Let us also take some generated function $u_b(x)$ and call it cardinal function. In this case we can construct Kunchenko's polynomial from generated functions $u_v(x)$, $v \neq b$, such functions we'll call additional functions:

$$P_r(x) \equiv \sum_{v=0, v \neq b}^{r} \alpha_v u_v(x) = \alpha_0 u_0(x) + \sum_{v=1, v \neq b}^{r} \alpha_v u_v(x). \quad (4)$$

Considering distance minimization between Kunchenko's polynomial (4) and cardinal function $u_b(x)$:

$$\rho_{bP}^2 = \int_a^d [u_b(x) - P_r(x)]^2 dx, \quad (5)$$

As shown in [12, p. 75-76] coefficients $\alpha_v$, $v \neq 0$ can be found as solution of following linear equations system:

$$\sum_{k=1, k \neq b}^{r} \alpha_k F_{v,k} = F_{v,b}, \quad v = \overline{1, r}, v \neq b, \quad (6)$$

where centered correlants are $F_{v,k} \equiv \Psi_{v,k} - \Psi_v \cdot \Psi_k \cdot \|u_0(x)\|^2$,

$$\Psi_v \equiv \frac{\Psi_{v,0}}{\|u_0(x)\|^2} = \int_a^d u_v(x) \cdot u_0(x) dx / \int_a^d u_0^2(x) dx.$$

Coefficient $\alpha_0$ must be equal to next expression:

$$\alpha_0 = \Psi_b - \sum_{v=1, v \neq b}^{r} \alpha_v \Psi_v. \quad (7)$$

Using coefficients (6), (7) in (5) we'll get:

$$\rho_{bP}^2 = F_{b,b} - J_r, \quad (8)$$

where $J_r$ is so-called polynomial's (4) inforkune that looks as follows:

$$J_r \equiv \sum_{v=1, v \neq b}^{r} \alpha_v F_{v,b} = \sum_{v=1, v \neq b}^{r} \sum_{k=1, k \neq b}^{r} \alpha_v \alpha_k F_{v,k}. \quad (9)$$

At last, let us introduce numerical factor of polynomial's (4) approximation of cardinal function $u_b(x)$:

$$e_r = J_r / \int_a^d (u_b(x) - \Psi_b \cdot u_0(x))^2 dx. \quad (10)$$

Coefficient (10) we'll call efficiency coefficient of additional functions' polynomial approximation to cardinal function. The closer to 1 ratio means better polynomial's approximation to cardinal function [12, p. 86].

It should be noted that the set of additional functions consists of totally or partially nonorthogonal functions and the cardinal function is not included to additional functions' set. Thus, efficiency coefficient (10) can be equal to 1 only in ultimate case when polynomial's degree $r$ approaches infinity.

## IV. EXPERIMENTAL RESULTS

To illustrate how this method works we have produced following example: input signal consists of 10 randomly generated Gaussians (see Fig. 1):

$$signal(x) = \sum_{i=1}^{10} \frac{1}{\sqrt{2\pi}\sigma_i} e^{-(x-\mu_i)^2/2\sigma_i^2},$$

where $\mu_1 = 0$, $\mu_k = \mu_{k-1} + 4(\sigma_{k-1} + \sigma_k), k = \overline{2,10}$,

$\sigma_i \sim Rand(0.2; 2), i = \overline{1,10}$.

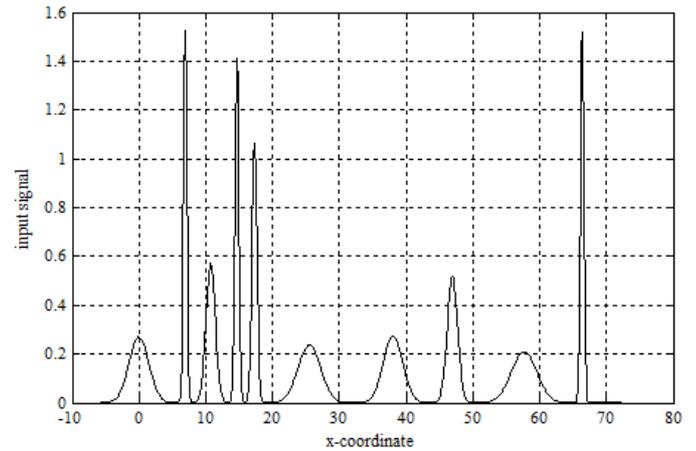

Figure 1. Input Signal (example)

As template, simple Gaussian was used:

$$template = \frac{1}{\sqrt{2\pi}} e^{-x^2/2}, -3 \leq x \leq 3.$$

We also used cross-correlation (CC) and sum of squared differences (SSD) to compare efficiency of our approach with these two methods.

Graphs of efficiency we'll call effectograms. They are shown on Fig. 2 (for all three methods).

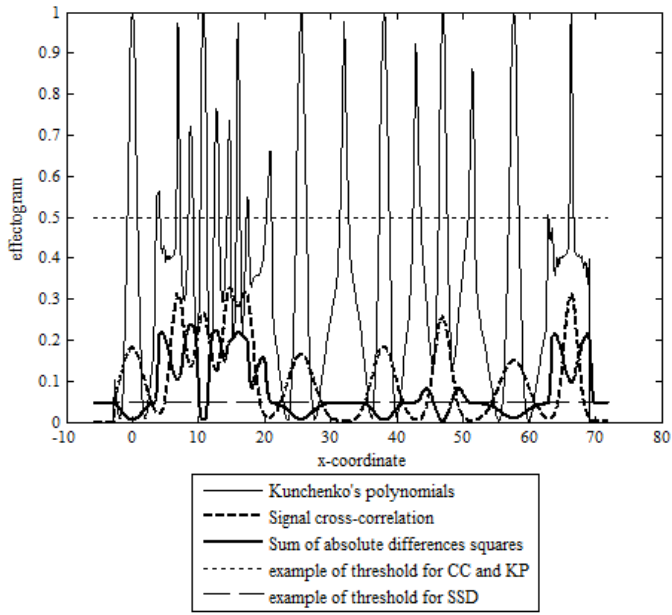

Figure 2. Effectograms of Kunchenko's polynomials, cross-correlation and sum of squared differences methods

The effectograms plot shows that introduced method gives the best matching results – its efficiency is much closer to 1 than two other methods.

However, there is one interesting detail about Kunchenko's polynomials method (KP). This method also finds "inverted" templates (Gaussians in our example). You can see effectogram peaks with *x*-coordinates that correspond to probable inverted Gaussians in input signal. It can be considered both as advantage and as disadvantage.

Result of statistical experiment presented in Table I. The quantity of experiment's sample is 100.

TABLE I. RESULT OF STATISTICAL EXPERIMENT

| Kunchenko's polynomials | | | | | |
|---|---|---|---|---|---|
| *Threshold level* | 0.4 | 0.5 | 0.6 | 0.7 | 0.8 |
| *Found Gaussians* | 8.46 | 9.55 | 8.39 | 7.59 | 6.96 |
| *Wrong Gaussians* | 1.77 | 0.39 | 1.17 | 1.67 | 1.79 |
| Cross-correlation | | | | | |
| *Threshold level* | 0.4 | 0.5 | 0.6 | 0.7 | 0.8 |
| *Found Gaussians* | 0.22 | 0 | 0 | 0 | 0 |
| *Wrong Gaussians* | 0.02 | 0 | 0 | 0 | 0 |
| Sum of squared differences | | | | | |
| *Threshold level* | 0.05 | 0.1 | 0.15 | 0.2 | 0.25 |
| *Found Gaussians* | 7.38 | 4.12 | 2.82 | 1.96 | 1.31 |
| *Wrong Gaussians* | 1.23 | 0.54 | 0.44 | 0.28 | 0.13 |

Generalized statistical experiment's algorithm looks as follows:
- Thresholding effectograms. We used two different thresholds for KP and CC methods and another one for SSD (see Fig. 2)
- Searching for local maximums in KP and CC thresholded effectograms and minimums in SSD one.
- Comparing found maximums/minimums with trusted intervals of real means in input signal. If local maximum/minimum occurred in trusted interval $[\mu_i - \delta; \mu_i - \delta]$, $\overline{i = 1,10}$, we assume that Gaussian has been found. Maximum derivation from means was established experimentally and equals to $\delta = 0.1$.

## V. CONCLUSION AND FUTURE RESEARCH

Statistical experiment illustrates that template matching method based on KP has several advantages over classical methods (CC and SSD): comparing to CC, KP makes it possible to find templates in input signal with higher threshold level. As of SSD – introduced method fetches more Gaussians when high confidence level is used.

As mentioned, KP has particular feature – "inverted" templates searching. On the one hand, this feature can be considered as possibility of distorted templates searching and, on the other hand, KP used as classical template matching method for 1D signal processing should be improved to eliminate "inverted" templates phenomena.

REFERENCES

[1] R. Brunelli, Template Matching Techniques in Computer Vision: Theory and Practice. Chippenham, UK: John Wiley & Sons, 2009.
[2] Y.-H. Lin and C.-H. Chen, "Template matching using the parametric template vector with translation, rotation and scale invariance," Pattern Recognition, vol. 41, pp. 2413-2421, July 2008.
[3] W. A. Fellenz, J. G. Taylor, N. Tsapatsoulis, and S. Kollias, "Comparing Template-based, Feature-based and Supervised Classification of Facial Expressions from Static Images," Proc. of "Multiconference on Circuits, Systems, Communications and Computers," pp. 5331-5336, 1999.
[4] S. Haykin, Neural Networks: A Comprehensive Foundation, 2nd ed. Upper Saddle River, NJ: Prentice Hall, 1998.
[5] A. Margalit, and A. Rosenfeld, "Using feature probabilities to reduce the expected computational cost of template matching," Computer Vision, Graphics and Image Processing, vol. 52, pp. 110-123, 1990.
[6] K. Fredriksson, G. Navarro, and E. Ukkonen, "Sequential and indexed two-dimensional combinatorial template matching allowing rotations," Theoretical Computer Science, vol. 347, pp. 239-275, 2005.
[7] A. Rosenfeld, and A. C. Kak, Digital Picture Processing, 2nd ed. NY: Academic Press, 1976. (Comp. Science and Appl. Mathematics series).
[8] D.-M. Tsai, and C.-T. Lin, "Fast normalized cross correlation for defect detection," Pattern Recognition Letters, vol. 24, № 15, pp. 2625-2631, 2003.
[9] J. P. Lewis, "Fast Template Matching," Proceedings of Conference "Vision Interface 95," pp. 120-123, 1995.
[10] O. Chertov, D. Tavrov, D. Pavlov et al., Group methods of data processing. Raleigh, NC: Lulu.com, 2010.
[11] Y. P. Kunchenko, Polynomial parameter estimations of close to Gaussian random variables. Aachen: Shaker, 2002.
[12] Y. P. Kunchenko. Approximations in a Space with Generative Element. Kyiv: Naukova Dumka, 2003. (in Russian)